\documentclass[pdflatex,sn-mathphys-num]{sn-jnl}

\usepackage{graphicx}%
\usepackage{multirow}%
\usepackage{amsmath,amssymb,amsfonts}%
\usepackage{amsthm}%
\usepackage{mathrsfs}%
\usepackage[title]{appendix}%
\usepackage{xcolor}%
\usepackage{textcomp}%
\usepackage{manyfoot}%
\usepackage{booktabs}%
\usepackage{algorithm}%
\usepackage{algorithmicx}%
\usepackage{algpseudocode}%
\usepackage{listings}%

\usepackage{natbib}
\setcitestyle{numbers,square}
\usepackage{bbm}
\usepackage{amsmath}
\usepackage{subcaption} 
\usepackage{booktabs}

\usepackage[normalem]{ulem}
\usepackage{xcolor}

\usepackage{mdframed}
\usepackage{caption}

\theoremstyle{thmstyleone}%
\theoremstyle{thmstyletwo}%

\theoremstyle{thmstylethree}%

\begin{document}

\title[Article Title]{Beyond Visual Cues: Leveraging General Semantics as Support for Few-Shot Segmentation}


\author[1]{\fnm{Jin Wang} }\email{wangjin@s.upc.edu.cn;  liumengyv@s.upc.edu.cn; Jian.Pang@geely.com; chenhl@upc.edu.cn}

\author*[1]{\fnm{Bingfeng Zhang}}\email{liuwf@upc.edu.cn; bingfeng.zhang@upc.edu.cn}

\author[2]{\fnm{Jian Pang}}
\author[1]{\fnm{Mengyu Liu}}
\author[1]{\fnm{Honglong Chen}}
\author*[1]{\fnm{Weifeng Liu}}

\affil[1]{\orgdiv{School of Control Science and Engineering}, \orgname{China University of Petroleum (East China)}, \orgaddress{\street{66 West Changjiang Road}, \city{Qingdao}, \postcode{266580}, \state{Shandong}, \country{China}}}
\affil[2]{\orgdiv{Geely Automobile Research Institute}, \orgname{Zhejiang Geely Holding Group Co., Ltd.}, \orgaddress{\street{818 Binhai 2nd Road}, \city{Ningbo}, \postcode{315000}, \state{Zhejiang}, \country{China}}}



\abstract{Few-shot segmentation (FSS) aims to segment novel classes under the guidance of limited support samples by a meta-learning paradigm. Existing methods mainly mine references from support images as meta guidance. However, due to intra-class variations among visual representations, the meta information extracted from support images cannot produce accurate guidance to segment untrained classes. In this paper, we argue that the references from support images may not be essential, the key to the support role is to provide unbiased meta guidance for both trained and untrained classes. We then introduce a Language-Driven Attribute Generalization (LDAG) architecture to utilize inherent target property language descriptions to build robust support strategy. Specifically, to obtain an unbiased support representation, we design a Multi-attribute Enhancement (MaE) module, which produces multiple detailed attribute descriptions of the target class through Large Language Models (LLMs), and then builds refined visual-text prior guidance utilizing multi-modal matching. Meanwhile, due to text-vision modal shift, attribute text struggles to promote visual feature representation, we design a Multi-modal Attribute Alignment (MaA) to achieve cross-modal interaction between attribute texts and visual feature. Experiments show that our proposed method outperforms existing approaches by a clear margin and achieves the new state-of-the art performance. The code will be released.}




\maketitle

\section{Introduction}\label{sec1}

With the development of deep learning~\cite{lecun2015deep, vaswani2017attention} and foundation models~\cite{lin2023clip, roy2023sam, radford2021learning}, semantic segmentation~\cite{jiao2023learning, minaee2021image} has made great progress. Semantic segmentation models need massive pixel-level labels. They perform poorly on unseen classes due to weak generalization. This further limits the use in open environments. To solve the above problems, few-shot segmentation~\cite{fan2022self, lang2023base, long2015fully, min2021hypercorrelation, peng2023hierarchical, zhang2021self, zhang2019pyramid} (FSS) is proposed, which inspires by the human ability to learn new things quickly; FSS aims to perform segmentation on novel classes based on the knowledge learned from few annotated seen classes. To achieve generalization to novel classes, FSS develops a meta-learning paradigm, which first divides the data into a support set and a query set, and then extracts the meta-knowledge from the support set to segment the images in the query set. 

\begin{figure*}[!t]
	\centering
	\includegraphics[width=1.0\textwidth]{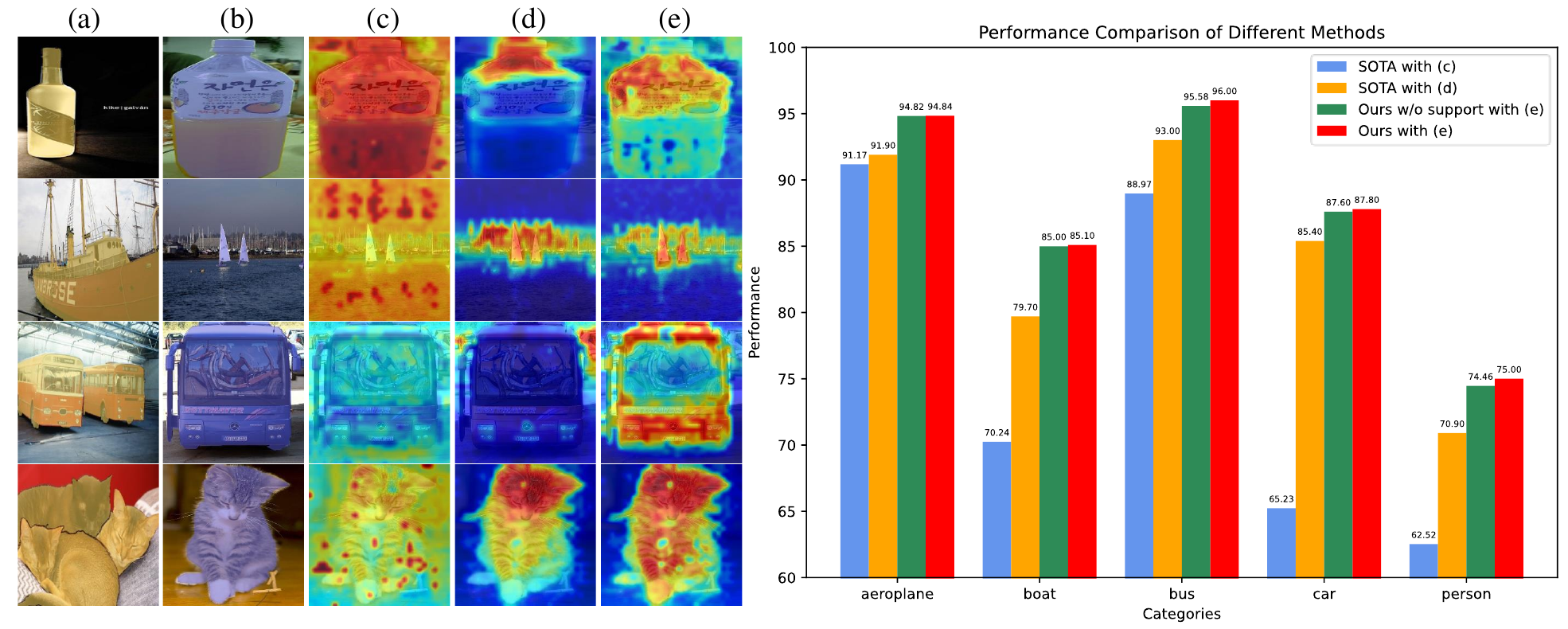}
	\caption{Comparison of prior masks and performance on some common classes using different FSS methods. (a) Support images with ground-truth masks (yellow); (b) Query images with ground-truth masks (blue); (c) Prior information from previous visual-visual matching SOTA method~\cite{sun2024vrp}, which is difficult to capture target class; (d) Prior information from previous visual-text alignment SOTA method~\cite{wang2024rethinking} with fixed text template ``a photo of \{\emph{target class}\}'', which focus only on parts of target class, \emph{e.g.,} the mirror of bus, the head of cat. (e) Our proposed multiple visual-text prior information, which focuses more on the overall area of target class rather than locally distinct areas.}
	\label{fig_motivation}
\end{figure*}

The previous FSS model mainly concentrates on mining meta information from the support image to segment the query images. Depending on how the support information is utilized, existing methods can be classified into two categories: prototype matching methods~\cite{luo2023pfenet++, zhang2021self, tian2020prior, peng2023hierarchical, xu2024eliminating} and prior instruction methods~\cite{chen2024visual, hong2022cost, lang2023base, sun2024vrp, wang2024rethinking}. Prototype matching methods extract prototypes from support images for similarity calculation or dense matching with the query image. Prior instruction methods produce an initial query target localization to guide the segmentation by using the support-query image pixel-level correspondence. Whether prototype matching or prior instruction methods, they rely heavily on support images to extract references or guidance for building support-query matching. However, this support-query matching paradigm results in inaccurate target identification, as shown in Fig.~\ref{fig_motivation} (c), and suffers from following problems: 1) the inherent intra-class visual differences between support and query images, which prevent the support images from providing reliable references; 2) the bias to the trained classes, which excessive focus on trained classes rather hinders the learning of untrained classes and consequently limiting the generalization capability of FSS models.

Based on the above analysis, we posit that effective support information should satisfy two fundamental requirements: 1) category-agnostic property (\emph{i.e.}, independence from both trained and untrained classes), and 2) the capacity to provide discriminative target information. Current methods heavily rely on support images to produce support guidance, easily failing to meet these criteria. We argue that the key to FSS does not primarily depend on the support images themselves, but rather on providing robust references that facilitates effective matching with query images. Thus, we attempt to utilize text in place of support images. Compared to visual data, textual information naturally possesses stronger generalization capabilities independent of category interference. In addition, the textual presentation is more flexible, allowing describe the target class from different perspectives. For example, the bus is \emph{square in shape}, \emph{with large glass windows around the body} and \emph{large reflectors on both sides}. These three inherent semantic attributes are robust and can help the model quickly identify the bus with only a small number of samples, without having to learn these features from a large amount of visual data.
As shown in Fig.~\ref{fig_motivation}, we conduct experiments comparing the performance of support information from images and text, respectively. We found that visual-text matching responses (Fig.~\ref{fig_motivation}(d)) are more effective than visual-visual matching responses (Fig.~\ref{fig_motivation}(c)), indicating that visual-text support information is more reliable than visual-visual support information when only limited support samples are provided.

In this paper, we propose a Language-Driven Attribute Generalization (LDAG) method with multiple attribute text descriptions for FSS. Our LDAG is designed with two considerations: 
1) introducing class-consistent textual descriptions to compensate for the unreliable visual references caused by intra-class variations; 2) providing unbiased reference information to ensure consistent reference capabilities across both trained and untrained data in the FSS task. Specifically, we design two modules to introduce textual information efficiently. The first is Multi-attribute Enhancement (MaE), which acquires multiple valid category attribute text descriptions by conducting question-and-answer sessions with Large Language Models (LLMs) on the target category attribute knowledge. With multiple attribute details added, multiple fine-grained prior instructions can be generated and the focus on the target category is enhanced. The second is called Multi-modal Attribute Alignment (MaA), which employs contrastive learning to align visual global features with textual attribute embeddings. This text-visual attribute alignment enhances the expressive power of visual features with semantic attributes. MaE introduces fine-grained attributes extracted by LLMs as references, and MaA aligns text attributes to visual features through contrast learning, and the two synergistically enhance attribute semantic perception in FSS. As shown in Fig.~\ref{fig_motivation} (e), our proposed multiple attribute textual descriptions with strong generalization ability can focus more on the whole region of the target class rather than part of the region of interest (Fig.~\ref{fig_motivation} (d), which generate text-vision prior based on one fixed ``a photo of \{\emph{target class}\}'' template), enhancing the ability to recognize the target class and generalization to novel classes.

Our contributions are summarized as follows:
\begin{itemize}
\item We argue that support images provide limited reference information for meta-learning in FSS and propose a novel Language-Driven Attribute Generalization (LDAG) framework, which leverages LLM-derived attribute semantics to construct a more discriminative and generalizable reference representation for FSS tasks.
\item To introduce textual information as references for FSS tasks, we design Multi-attribute Enhancement (MaE) to obtain multiple pixel-level multi-modal guidance. Furthermore, a Multi-modal Attribute Alignment (MaA) is performed to obtain a more comprehensive visual feature representation for efficient decoding.
\item Our method has a significant improvement over existing methods on both PASCAL-5$^{i}$~\cite{shaban2017one} and COCO-20$^{i}$~\cite{nguyen2019feature} datasets and achieves state-of-the-art performance.
\end{itemize}

\section{Related Work}\label{sec2}
\subsection{Few-Shot Segmentation} 
Few-shot segmentation aims to transfer the knowledge learned from the small amount of labeled seen classes to the unseen classes and achieves the dense prediction to the novel classes. Most existing few-shot segmentation methods followed the idea of metric-based meta-learning~\cite{karimi2025dsv, snell2017prototypical, shi2022dense, zhang2022mask}.  

PFENet~\cite{tian2020prior} first proposes a training-free prior mask generation method based on the visual relationship between support images and query images, furthermore, it enriches the query feature with support features for decoding under the guidance of prior masks. PFENet++~\cite{luo2023pfenet++} improves on the existing prior mask strategy by using more contextual information to generate higher-quality prior masks. BAM~\cite{lang2022learning} designs an additional branch to generate the prior masks for seen classes to enhance the ability to discriminate between unseen classes. HDMNet~\cite{peng2023hierarchical} utilizes the transformer architecture for pixel-level support correlation as prior masks to alleviate the overfitting to base classes. AENet~\cite{xu2024eliminating} mines discriminative query foreground regions to rectify the ambiguous foreground features and achieves an efficient support-query matching for prior mask generation. RD~\cite{zhou2024unlocking} constructs a relationship descriptor as prior mask across multiple layers by utilizing the pre-trained vision transformers to enhance the generalization. PGMAM~\cite{chen2024visual}
proposes a class-agnostic mask assembly process to generate multi-modal prior mask to alleviate the bias to base classes. PI-CLIP~\cite{wang2024rethinking} rethinks the generation of prior masks and proposes to replace the original visual prior representation with the visual-text alignment capacity to capture more reliable guidance.

Existing methods rely on extracting reference information from support images for query matching, our approach introduces text attributes as reference and leverages foundation models to generate multiple robust text-query matching.

\subsection{Large Language Models} 
In recent years, Large Language Models (LLMs) have significantly impacted natural language processing with their powerful text generation capabilities. Pre-trained models like GPT-3~\cite{brown2020language} and PaLM~\cite{chowdhery2023palm}, based on the Transformer architecture~\cite{vaswani2017attention}, have achieved breakthroughs in open-domain text generation through self-supervised learning. These models trained on massive data, generate coherent and semantically rich text. Their ability to understand precise intentions via prompt fine-tuning and dynamically adapt through contextual reasoning enables the generation of target attribute information for quizzing LLMs.

\subsection{Vision Foundation Models} 
Vision foundation models like CLIP and SAM have significantly advanced the field of computer vision. CLIP enables powerful multi-modal understanding through its ability to align image and text representations, supporting applications such as zero-shot segmentation, object detection. SAM, on the other hand, excels in versatile segmentation tasks, including real-time video processing and interactive object segmentation.

The Segment Anything Model (SAM)~\cite{kirillov2023segment} consists of an image encoder, a prompt encoder, and a mask decoder. The image encoder extracts features using pre-trained weights, while the prompt encoder processes user inputs like points, boxes, and masks. The mask decoder then generates segmentation masks based on these inputs. Trained on over 1 billion masks from 11 million images, SAM captures rich visual information and demonstrates strong generalization, performing well in few-shot~\cite{zhang2023personalize, bai2024fs} and zero-shot learning tasks~\cite{li2024clipsam, roy2023sam, deng2023segment} with minimal task-specific data.

The Contrastive Language-Image Pretraining (CLIP) model~\cite{radford2021learning} aligns text and visual features using a text encoder and a visual encoder via contrastive learning. Trained on 400 million text-image pairs, CLIP integrates semantic information from both modalities, overcoming the limitations of traditional visual features. This enables CLIP to perform well in downstream applications such as object detection~\cite{ju2022adaptive}, semantic segmentation~\cite{lin2023clip, yang2023multi, shuai2023visual}, etc., making it a fundamental tool for advancing vision-language tasks.

\section{Method}\label{sec3}
 \subsection{Task Description}

Few-shot segmentation focuses on transferring segmentation knowledge from known base classes to unknown novel classes using limited labeled data. It typically follows the meta-learning framework. The dataset $D$ is split into a training set ${D_{train}}$ and a test set ${D_{test}}$ with disjoint class sets ${C_{train}}$ and ${C_{test}}$. Both ${D_{train}}$ and ${D_{test}}$ contain support set $S$ with image-mask pairs $\{I_s, M_s\}$ and a query set $Q$ with image-mask pairs $\{I_q, M_q\}$. During training, the model predicts segmentation masks for query images using the support set and is optimized based on the supervision of query masks. In inference, the model directly predicts the test set without further optimization, using query masks only for performance evaluation.

\subsection{Method Overview} 
\begin{figure*}[!t]
	\centering
	\includegraphics[width=0.95\textwidth]{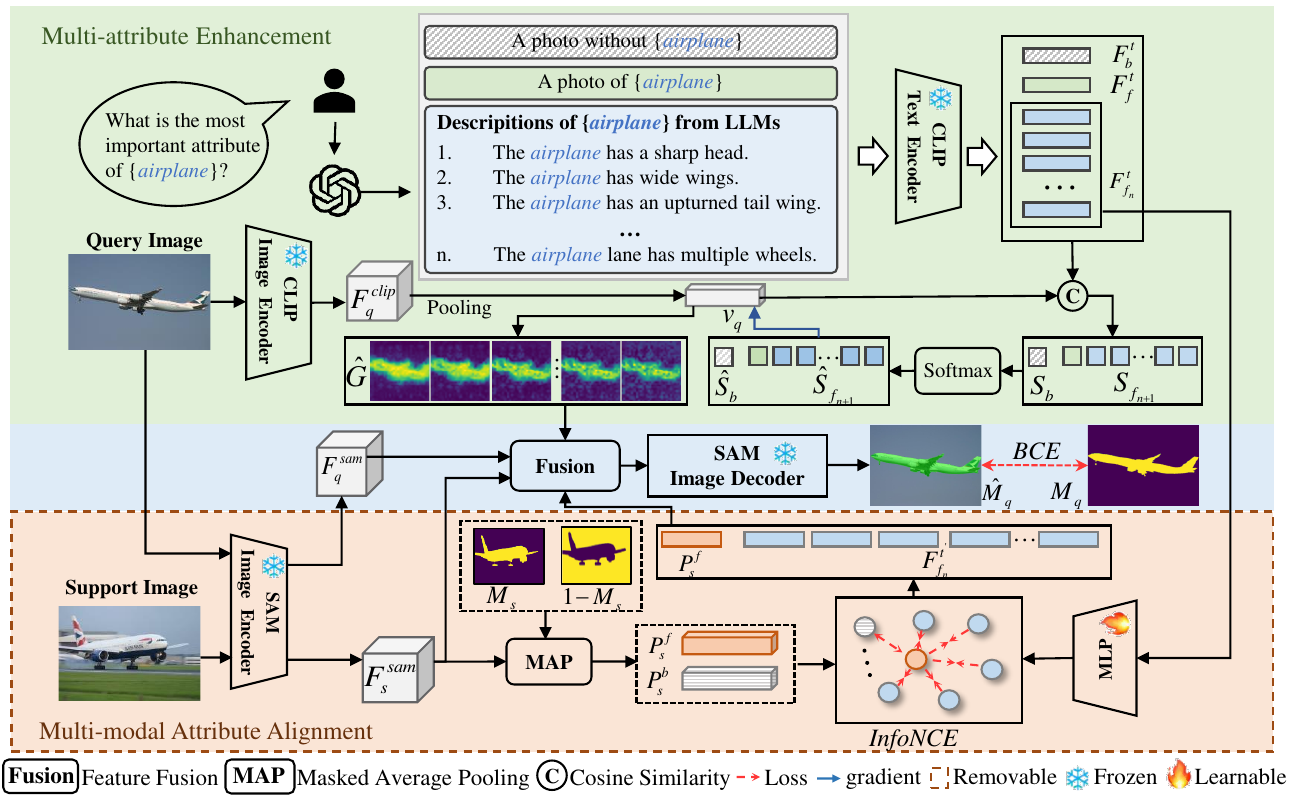}
	\caption{Overview of LDAG. MaE leverages LLMs to generate attribute text through Q$\&$A reasoning, multiple attribute prior is generated utilizing softmax-GradCAM. MaA promotes visual attribute representations via text-vision contrastive learning. Notably, competitive performance is preserved when removing MaA, \emph{i.e.}, support images are deleted, demonstrating its inherent robustness.}
	\label{fig_method}
\end{figure*}

In our LDAG, the frozen CLIP first generates multi-modal pixel-level prior instructions by exploiting its strong image-text alignment capability, which effectively captures target semantics and attributes. Then, for visual attribute decoding, we employ the segmentation capacity of SAM, which originates from the prompt-driven architecture and large-scale training, ensuring feature consistency and precise target localization.
Fig.~\ref{fig_method} shows our framework of the one-shot case with the following steps:
\begin{itemize}

    \item Firstly, a query image is sent to the frozen CLIP image encoder for image features, and its corresponding target class name is fed into the LLM to generate multiple attribute descriptions, which are then to generate text embedding with the frozen CLIP text encoder. By calculating image-text matching scores as gradient, multiple CAMs are obtained by Grad-CAM~\cite{selvaraju2017grad} in our proposed MaE module. 
    \item Then, the support and query images are encoded by frozen SAM encoder to extract query features and support prototypes. These prototypes serve as anchors and negative samples, while the attribute text embeddings act as positive samples for contrastive learning in our MaA module.
    \item Ultimately, the support foreground prototype, multiple attribute text embeddings, multiple CAMs, and query features are cascaded into the SAM decoder to obtain the final prediction. 
    
\end{itemize}

Note that we experimentally found that even removing support prototypes and MaA, \emph{i.e.}, support images are not used, the performance of our LDAG is also impressive. See Table~\ref{tab:supp} for more details.

\subsection{Multi-attribute Enhancement}
Current FSS methods extract reference information from the support image to guide the segmentation process, since the intra-class differences between the support image and the query image, the support image is difficult to provide efficient reference information and limits untrained target recognition. While replacing support images with textual context can provide effective reference information, relying solely on target class names, \emph{e.g.}, using ``a photo of [class]'', is insufficient for locating the target region accurately. As shown in Fig.~\ref{fig_motivation}(d), this simple textual representation fails to capture critical visual attributes and intra-class variations, ultimately limiting the generalization capability across diverse instances. Therefore, it is necessary to design complete and robust information to enhance the recognition ability of the target class. We propose to obtain pixel-level target indications by leveraging multiple attribute description capabilities of large language models.

Specifically, we adapt LLMs, \emph{e.g.}, GPT-o1, to generate descriptions of intrinsic category properties that do not vary with viewpoint, size, \emph{etc.} Given one target class $c$ from the dataset with $M$ class $C=\{c_1, c_2, ..., c_M\}$, we have designed the following instructions for the LLMs: \textbf{\emph{`` There are ${M}$ classes in a dataset: $c_1, c_2, ..., c_M$,  List n descriptions with key properties to describe the $c$ in terms of appearance, color, shape, size, or material, etc. These descriptions will help visually distinguish the $c$ from other classes in the dataset. Each description should follow the format: `a clean origami $c$. It + descriptive contexts'. Do not have any content output other than the given format. And try not to include any other class names in the description.''}} After that, by adding a fixed expression `` a photo of \{\emph{target class}\}'', we get a total of $n + 1$ target text descriptions. The non-target prompt is designed as ``a photo without \{\emph{target class}\}'' and they are sent to the CLIP text encoder to get the text embeddings, represented as foreground text embedding $F_{f_{n+1}}^{t}$ and background text embedding $F_{b}^{t}$.

Then, we use the above $n+1$ robust attribute representations to generate query instructions. Suppose the query image is $I_q$ and sent it to the CLIP image encoder to get the clip query features $F_q^{clip}$, after removing the class token and performing pooling of the $F_q^{clip}$, query token $v_q$ is obtained. Then, the foreground/background visual-text similarity scores are obtained by:
\begin{equation}
S_{f_i} = \frac{v_q^\mathrm{T}F^t_{f_i}}{|v_q||F^t_{f_i}|}/\tau, i \in \{ 1, 2, \ldots, n+1 \};   \qquad    S_b = \frac{v_q^\mathrm{T}F^t_b}{|v_q||F^t_b|}/\tau,
\end{equation}
where $\tau$ is a temperature parameter, $S_{f_i}$ represents the foreground scores and $S_b$ represents the background score, to further improve the focus on target areas and reduce the impact of background areas, a $softmax$ operation is performed:
\begin{equation}
\hat{S_{f_i}}, \hat{S_b}=softmax(S_{f_i}, S_b), i \in \{ 1, 2, \ldots, n+1 \},
\end{equation}
following Grad-CAM~\cite{selvaraju2017grad}, by calculating the gradient of the foreground similarity score $\hat{S_{f_i}}$ over $F_q^{clip}$, we can get weights $\omega$ of the foreground part, and prior guidance $G$ are obtained by:
\begin{equation}
G=ReLU(\sum_{m}^{} \omega_i(m) F_{q}^{clip}(m)), i \in \{ 1, 2, \ldots, n+1 \},
\end{equation}
where $m$ represents the index of feature map, $ReLU$ represents the ReLU activation function, and the refined prior guidance $\hat{G}$ are obtained following ~\cite{wang2024rethinking}, which is utilized as the target instruction in the final decoding process. By employing fine-grained prompt engineering to query LLMs, we obtain diverse expressions of the most discriminative attributes for the target category. This multiple attribute-aware text-vision alignment enables the generated refined prior instructions to identify broader target regions, effectively mitigating issues associated with limited areas of concern.

\subsection{Multi-modal Attribute Alignment}
The previously mentioned MaE extends the focus target regions by exploiting fixed attributes, but the decoding for FSS requires not only rich prior information about the target, but also discriminative visual feature representations. Due to the problem of text-vision modal variation, attribute text does not directly enhance the representational power of frozen visual features, limiting the use of target attribute information in decoding process. In order to get effective visual representations, we design a cross-modal attribute alignment strategy named Multi-modal Attribute Alignment (MaA), which further exploits robust attribute text and extends attribute representation capability to visual modal.

Specifically, given the support image $I_s$ and query image $I_q$, after passing through the frozen SAM image encoder, corresponding SAM image support feature $F_s^{sam}$ and query feature $F_q^{sam}$ are obtained. With the support mask $M_s$, support foreground prototype $P_s^f$ and background prototype $P_s^b$ are generated using the MAP~\cite{zhang2020sg, zhang2019canet, zhu2021self}:
\begin{equation}
P_s^f= \frac{\sum_{i=1}^{hw} F_{s}^{sam}(i) \cdot \mathbbm{1} (M_s(i)=1)}{\sum_{i=1}^{hw} \mathbbm{1} (M_s(i)=1)},    \qquad    P_s^b= \frac{\sum_{i=1}^{hw} F_{s}^{sam}(i) \cdot \mathbbm{1} (M_s(i)=0)}{\sum_{i=1}^{hw} \mathbbm{1} (M_s(i)=0)},
\end{equation}
where $i$ represents the pixel index, $h$ and $w$ represent the height and width of the feature map, respectively. $\mathbbm{1}$ is the indicator function, $M_s(i)=1$ indicates the $i$-th pixel belongs to class $c$, vice versa. Note that $M_s$ is reshaped as the same size with $F_s^{sam}$. 

To enhance the representation of textual features in visual space and thus optimize cross-modal alignment effects, we first transform each of the $n$ attribute text embeddings $F_{f_n}^t$ using a dedicated MLP projection:
\begin{equation}
F_{f_i}^{t'} = \text{MLP}_i(F_{f_i}^{t}), \quad i  \in \{ 1, 2, \dots, n \}
\end{equation}

Building upon these transformed features, we establish contrastive relationships between the foreground prototype $P_s^f$, background prototype $P_s^b$, and the attribute textual embeddings $F_{f_i}^{t'}$ through an InfoNCE loss:
\begin{equation}
\mathcal{L}_{\text{Inf}} = -\frac{1}{n} \sum_{i=1}^{n} \log \left( 
  \frac{
    \exp\left( \operatorname{sim}\left( P_s^f, F_{f_i}^{t'} \right) / \tau_1 \right)
  }{
    \exp\left( \operatorname{sim}\left( P_s^f, F_{f_i}^{t'} \right) / \tau_1 \right) + 
    \exp\left( \operatorname{sim}\left( P_s^f, P_s^b \right) / \tau_1 \right)
  } 
\right), i  \in \{ 1, 2, \dots, n \},
\end{equation}
where $\tau_1$ is another temperature parameter. To further enrich the visual feature representation with attribute information and create more robust references, we generate enhanced support features through feature fusion:
\begin{equation}
F_s^{sam'} = \mathcal{F}_1(F_s^{sam} \oplus F_{f_i}^{t'} \oplus P_s^f ), \quad i \in \{ 1, 2, \ldots, n \},
\end{equation}
the size of $F_{f_i}^{t'}$ and $P_s^{f}$ are reshaped to the same size as $F_s^{sam}$, where $\mathcal{F}_1$ means a light fusion network, $\oplus$ means cascade in the channel dimension, the enhanced support feature is then as references to perform efficient matching with the query feature for the final query prediction $\hat{M_q}$:
\begin{equation}
\hat{M_q}= D_{sam}(\mathcal{F}_2(F_s^{sam'} \oplus F_q^{sam} \oplus \hat{G})),
\end{equation}
where $D_{sam}$ means the frozen SAM decoder and $\mathcal{F}_2$ means another light fusion network. We employ contrastive learning to facilitate the transfer of information from the textual modality to the visual modality. This approach endows visual information with enhanced attribute expressiveness, thereby enriching the visual content and facilitating a more efficient decoding process.

Notably, the model maintains strong performance without support images and contrastive loss, as the learned multi-attribute text embeddings effectively map to visual space and provide high-quality references. This further confirms that in FSS, effective reference information is more crucial than support images themselves.

\subsection{Training loss}
We apply binary cross-entropy for the final prediction, Supervised by the query mask $M_q$, the prediction loss is computed by :
\begin{equation}
\mathcal{L}_{\text{pre}}  = \text{BCE}(\hat{M_q},M_q)
\end{equation}

Besides, in order to make more effective use of textual attribute information to enhance the learning of visual attribute features, we introduced a balancing parameter $\alpha$ in the overall loss function to optimize the weight of the contrast loss:
\begin{equation}
\mathcal{L} = \mathcal{L}_{\text{pre}} + \alpha \mathcal{L}_{\text{Inf}}
\end{equation}

\section{Experiments}\label{sec4}

\subsection{Datasets and Evaluation Metrics}
To evaluate our approach, we employed the PASCAL-5$^{i}$~\cite{shaban2017one} and COCO-20$^{i}$~\cite{nguyen2019feature} datasets. The PASCAL-5$^{i}$~\cite{shaban2017one} is derived from the PASCAL VOC 2012~\cite{everingham2010pascal} and enhanced with SDS~\cite{hariharan2011semantic}, contains 20 object classes such as people, cars, and cats. The COCO-20$^{i}$~\cite{nguyen2019feature} dataset, based on MSCOCO~\cite{lin2014microsoft}, comprises over 120,000 images spanning 80 categories and presents a more complex challenge. We assess our method using mean intersection-over-union (mIoU) and foreground-background IoU (FB-IoU), following prior research~\cite{lang2022learning, peng2023hierarchical, tian2020prior}.

\subsection{Implementation details}
In all experiments, the images are set to 512$\times$512 pixels, the batch size is set to 8, the learning rate is set to $1\times10^{-4}$, $\alpha$=0.5, $n$=5, $\tau$=$\tau_1$=1. Following the baseline, the model is trained for 50 epochs using distributed training on two NVIDIA 3090 GPUs with both datasets. The CLIP pre-trained model is ViT-B-16~\cite{radford2021learning} and the SAM pre-trained model is ViT-h~\cite{kirillov2023segment}. For the 5-shot case, the final segmentation mask is generated by aggregating the predictions from five different support sets. Other settings such as image enhancement methods, optimizers, \emph{etc.}, are the same as before~\cite{sun2024vrp}, since baseline VRP-SAM did not show 5-shot results, for fair comparison, we reproduce it and obtain new 1-shot and 5-shot performance.

\begin{table*}[ht]
\caption{Performance comparisons with mIoU (\%) as a metric on PASCAL-5$^{i}$ dataset, the \textbf{bold} indicates the optimal performance. $*$ means the reproduce performance for the fair comparison.}
\label{tab:tab1}
\centering
\resizebox{\textwidth}{!}{%
\begin{tabular}{llccccc|ccccc}
\toprule
\multirow{2}{*}{Backbone} & \multirow{2}{*}{Method} & \multicolumn{5}{c|}{1-shot} & \multicolumn{5}{c}{5-shot} \\ 
\cmidrule(lr){3-7} \cmidrule(lr){8-12}
& & Fold0 & Fold1 & Fold2 & Fold3 & Mean & Fold0 & Fold1 & Fold2 & Fold3 & Mean \\
\midrule
resnet50 & PFENet (TPAMI'20)~\cite{tian2020prior} & 61.7 & 69.5 & 55.4 & 56.3 & 60.8 & 63.1 & 70.7 & 55.8 & 57.9 & 61.9 \\
resnet50 & SCL (CVPR'21)~\cite{zhang2021self} & 63.0 & 70.0 & 56.5 & 57.7 & 61.8 & 64.5 & 70.9 & 57.3 & 58.7 & 62.9 \\
resnet50 & HDMNet (CVPR'23)~\cite{peng2023hierarchical} & 71.0 & 75.4 & 68.9 & 62.1 & 69.4 & 71.3 & 76.2 & 71.3 & 68.5 & 71.8 \\
resnet50 & BAM (TPAMI'23)~\cite{lang2022learning} & 69.9 & 75.4 & 67.0 & 62.1 & 68.6 & 72.6 & 77.1 & 70.7 & 69.8 & 72.6 \\
resnet50 & HMNet (NeurIPS'24)~\cite{xu2024hybrid} & 72.2 & 75.4 & 70.0 & 63.9 & 70.4 & 74.2 & 77.3 & 74.1 & 70.9 & 74.1 \\
resnet50 & AENet(ECCV'24)~\cite{xu2024eliminating} & 71.3 & 75.9 & 68.6 & 65.4 & 70.3 & 73.9 & 77.8 & 73.3 & 72.0 & 74.2 \\
resnet50 & VRP-SAM$^{*}$ (CVPR'24)~\cite{sun2024vrp} & 73.9 & 78.3 & 70.6 & 65.0 & 71.9 & 76.3 & 76.8 & 69.5 & 63.1 & 71.4 \\
resnet50 & PI-CLIP (CVPR'24)~\cite{wang2024rethinking} & 76.4 & 83.5 & 74.7 & 72.8 & 76.8 & 76.7 & 83.8 & 75.2 & 73.2 & 77.2 \\
resnet50 & FCP (AAAI'25)~\cite{park2025foreground} & 74.9 & 77.4 & 71.8 & 68.8 & 73.2 & 77.2 & 78.8 & 72.2 & 67.7 & 74.0 \\
\midrule
ViT-B/16 & PAT (TPAMI'24)~\cite{bi2024prompt} & 68.3 & 73.2 & 66.2 & 60.1 & 67.0 & 73.3 & 77.6 & 75.1 & 69.5 & 73.9 \\
ViT-B/16 & \textbf{ours}-LDAG (VRP-SAM) & \textbf{80.4} & \textbf{85.4} & \textbf{74.8} & \textbf{75.6} & \textbf{79.0} & \textbf{80.5} & \textbf{85.4} & \textbf{75.2} & \textbf{75.6} & \textbf{79.2} \\
\bottomrule
\end{tabular}%
}
\end{table*}

\begin{table*}[ht]
\caption{Performance comparisons on COCO-20$^{i}$ dataset, the \textbf{bold} indicates the optimal performance. $*$ means the reproduce performance for the fair comparison.}
\label{tab:tab2}
\centering
\resizebox{\textwidth}{!}{%
\begin{tabular}{llccccc|ccccc}
\toprule
\multirow{2}{*}{Backbone} & \multirow{2}{*}{Method} & \multicolumn{5}{c|}{1-shot} & \multicolumn{5}{c}{5-shot} \\ 
\cmidrule(lr){3-7} \cmidrule(lr){8-12}
& &Fold0 &Fold1 &Fold2 &Fold3 &Mean &Fold0 &Fold1 &Fold2 &Fold3 &Mean\\
\hline
resnet50 & PFENet (TPAMI'20)~\cite{tian2020prior} & 34.3 & 33.0 & 32.3 & 30.1 & 32.4 & 38.5 & 38.6 & 38.2 & 34.3 & 37.4 \\
resnet50 & SCL (CVPR'21)~\cite{zhang2021self}   & 36.4 & 38.6 & 37.5 & 35.4 & 37.0 & 38.9 & 40.5 & 41.5 & 38.7 & 39.9 \\
resnet50 & HDMNet (CVPR'23)~\cite{peng2023hierarchical}& 43.8 & 55.3 & 51.6 & 49.4 & 50.0 & 50.6 & 61.6 & 55.7 & 56.0 & 56.0 \\
resnet50 & BAM (TPAMI'23)~\cite{lang2023base} &45.2  &55.1  &48.7  &45.0  &48.5  &48.3  &58.4  &52.7  &51.4  &52.7  \\ 
resnet50 & HMNet (NeurIPS'24)~\cite{xu2024hybrid} &45.5 & 58.7 & 52.9 &51.4 &52.1 &53.4 &64.6 &60.8 &56.8 &58.9 \\
resnet50 & AENet(ECCV'24)~\cite{xu2024eliminating}& 45.4 & 57.1 & 52.6 & 50.0 & 51.3 & 52.7 & 62.6 & 56.8 & 56.1 & 57.1 \\
resnet50 & VRP-SAM$^{*}$ (CVPR'24)~\cite{sun2024vrp}  & 44.3 & 54.3 & 52.3 & 50.0 & 50.2 & 50.5 & 59.5 & 56.9 &54.9 & 55.5 \\
resnet50 & PI-CLIP (CVPR'24)~\cite{wang2024rethinking} &49.3 &65.7 &55.8 & 56.3 &56.8 &56.4 &66.2 &55.9 &58.0 & 59.1  \\
resnet50 & FCP (AAAI'25)~\cite{park2025foreground}  & 46.4 & 56.4 & 55.3 & 51.8 & 52.5 & 52.6 & 63.3 & 59.8 & 56.1 & 58.0 \\
\hline
ViT-B/16 & PAT (TPAMI'24)~\cite{bi2024prompt} & 40.6 & 51.9 & 49.0 & 50.7 &48.0 & 51.2 & 63.2 & 59.5 & 55.9 &57.4 \\
ViT-B/16 &\textbf{ours}-LDAG (VRP-SAM)  & \textbf{55.4} & \textbf{64.9} & \textbf{61.6} & \textbf{60.8} &\textbf{60.5} &\textbf{56.4} &\textbf{66.2} &{\textbf{61.8}} & \textbf{62.4} &\textbf{61.7}  \\
\bottomrule
\end{tabular}%
}
\end{table*}

\begin{figure*}[!t]
	\centering
	\includegraphics[width=1.0\textwidth]{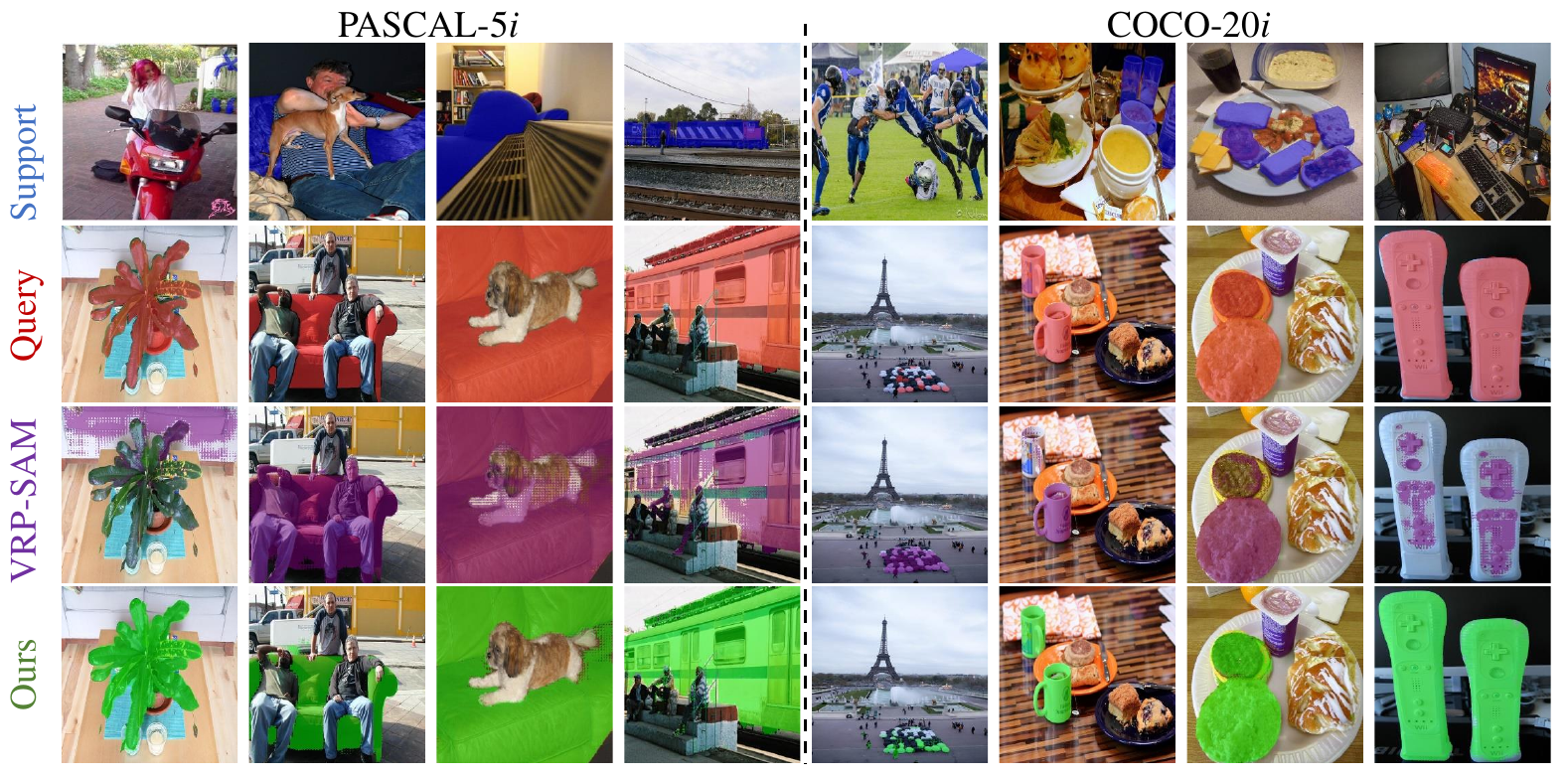}
	\caption{Qualitative results of the proposed LDAG and other SAM-based method (VRP-SAM~\cite{sun2024vrp}) approach under 1-shot setting. Each row from top to bottom represents the support images with ground-truth  (GT) masks (blue), query images with GT masks (red), baseline results (purple), and our results (green), respectively.}
	\label{fig_vis_final}
\end{figure*}

\subsection{Comparison with state-of-the-art}

\textbf{Quantitative results}
Table~\ref{tab:tab1} shows the performance of our method and existing state-of-the-art methods for few-shot segmentation on PASCAL-5$^{i}$, our approach greatly improves the performance of the model over the 1-shot and 5-shot tasks compared to different methods and achieves new state-of-the-art performance. For the CLIP-based PI-CLIP, our method increases mIoU with 2.2$\%$ for 1-shot and 2.0$\%$ for 5-shot. For the SAM-based methods, our approach outperforms other approaches by a clear margin, with mIoU gain of 7.1$\%$ for VRP-SAM and 5.8$\%$ for FCP on 1-shot, 7.8$\%$ for VRP-SAM and 5.2$\%$ for FCP on 5-shot.

In Table~\ref{tab:tab2}, we compare the performance of our approach and others on COCO-20$^{i}$ dataset. Our approach also exhibits strong performance and achieves new
state-of-the-art performance. Specifically, our approach improves the baseline VRP-SAM by 10.3$\%$ and 6.2$\%$ mIoU for 1-shot and 5-shot tasks.

\textbf{Qualitative results.}
To show the effect of our proposed model on the existing methods, we visualize the results of the VRP-SAM and our proposed method in Fig.~\ref{fig_vis_final}, it can be found that our method (green part) has a much stronger target localization ability than VRP-SAM (purple part), and the area of interest in the target category has expanded significantly.

\begin{figure*}[!t]
	\centering
	\includegraphics[width=1.0\textwidth]{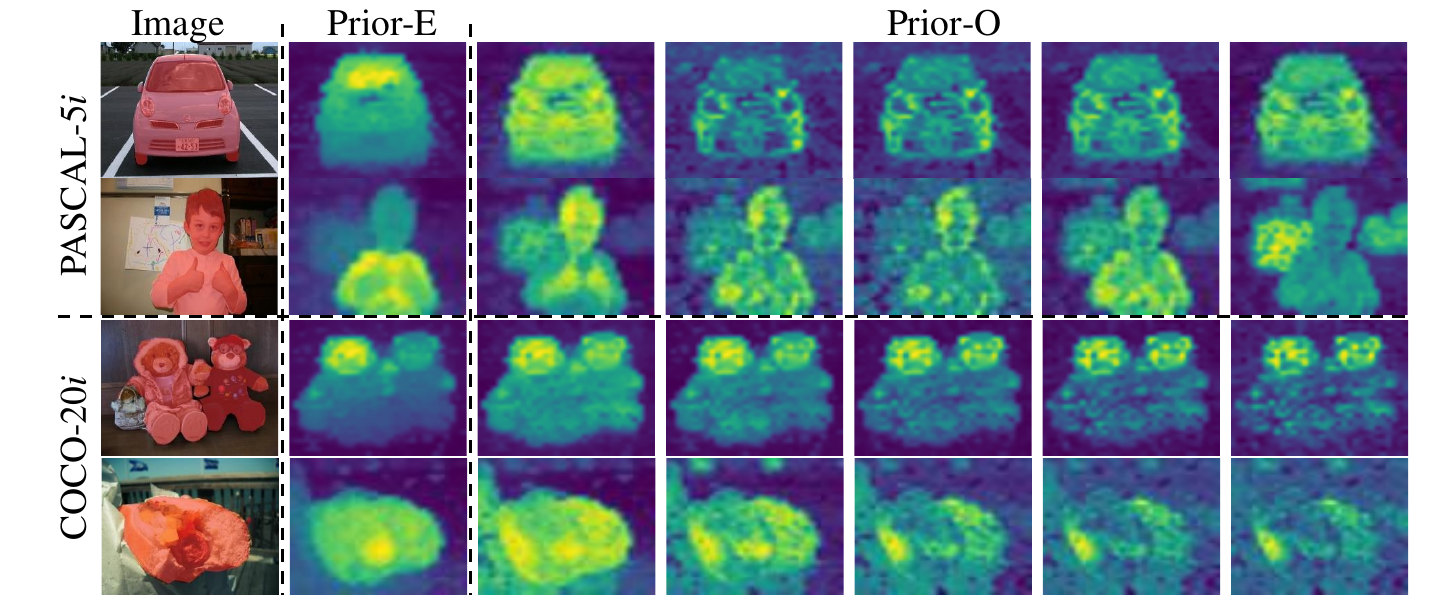}
	\caption{Visualization of prior information comparison from PASCAL-5$^{i}$~\cite{shaban2017one} and COCO-20$^{i}$~\cite{nguyen2019feature}. `` Image '' represents the query image, `` Prior-E '' represents the prior information generated by fixed text descriptions, `` Prior-O '' represents our proposed prior information. Our proposed method is able to focus on different regions of the target class through different attribute information, which provides richer reference information and reduces the dependence of the FSS task on support images.}
	\label{fig_vis_cam}
\end{figure*}

Fig.~\ref{fig_vis_cam} shows the visualization of our proposed prior information (Prior-O) and existing prior information (Prior-E) to help understand the broader localization capabilities. Our proposed method is able to focus on different regions of the target class through different attribute information, which provides abundant reference information and reduces the dependence of the FSS task on the support images.

\textbf{Computation Cost.}
Our approach greatly improves model performance while reducing resource consumption, mainly due to the frozen CLIP and SAM. We compare the GPU memory and inference time in Table.~\ref{tab_GPU} with the baseline.
\begin{table}[!h]
\caption{Computation costs comparision, all experiments are conducted on NVIDIA RTX™ 3090, and ``Inf Time" meas the inference time for one image.}
\centering
\begin{tabular}{llccc}
\hline
{Method} &{GPUs} &{Inf Time}  &{mIoU} &{FB-IoU}\\
\hline
HDMNet &11.7GB &0.16s &69.4 &80.6 \\
PI-CLIP &13.0GB &0.25s &76.4 &87.6 \\
VRP-SAM &19.2GB &0.16s &73.9 &84.6 \\
ours-LDAG &\textbf{5.6GB} &\textbf{0.13s} &\textbf{80.4} &\textbf{89.5} \\
\hline
\vspace{-2em}
\label{tab_GPU}
\end{tabular}
\end{table}

\subsection{Ablation Study}
All ablation studies are performed on PASCAL-5$i$ dataset to evaluate the contribution of each module.

\textbf{Ablation Study on MaE and MaA.}
Efficient target prior and reference information are intuitively important for FSS, so we conduct relevant ablation studies on MaE and MaA. As can be seen in Table~\ref{tab:ablation_modules}, MaE yields a performance improvement of 5.7$\%$ and MaA yields a performance improvement of 0.8$\%$.

\textbf{Ablation Study on balancing parameter $\alpha$.}
Table~\ref{tab:alpha_ablation} shows the influence of the proportion a of predicted loss $\mathcal{L}_{\text{pre}}$ and contrastive loss $\mathcal{L}_{\text{Inf}}$ on the experimental results. When $\alpha$ is 0.5, the model effect reaches the best 80.40$\%$.

\textbf{Ablation Study on number $n$ of text descriptions}
Table~\ref{tab_ablation_number} shows the influence of the number $n$ of different attribute text descriptions on the experimental results. When $n=5$, the model effect reaches the best 80.40$\%$, and as $n$ increases, it will no longer contribute much to the growth of the model performance.

\begin{table}[h]
\centering
\begin{minipage}[t]{0.48\textwidth} 
\centering
\caption{Ablation study of MaE and MaA.}
\label{tab:ablation_modules}
\begin{tabular}{@{}c@{\hspace{0.5em}}c@{\hspace{0.5em}}c@{\hspace{0.5em}}c@{\hspace{0.5em}}c@{}}
\toprule
baseline & MaE & MaA & mIoU (\%) & FB-IoU (\%) \\
\midrule
\checkmark & & & 73.90 & 85.43 \\
\checkmark & \checkmark & & 79.60 & 89.02 \\
\checkmark & \checkmark & \checkmark & \textbf{80.40} & \textbf{89.52} \\
\bottomrule
\end{tabular}
\end{minipage}
\hfill
\begin{minipage}[t]{0.48\textwidth}  
\centering
\caption{Ablation study on parameter $\alpha$.}
\label{tab:alpha_ablation}
\begin{tabular}{@{}lcc@{}}
\toprule
$\alpha$ & mIoU (\%) & FB-IoU (\%) \\
\midrule
0 & 78.98 & 88.63 \\
0.3 & 79.98 & 89.28 \\
0.5 & \textbf{80.40} & \textbf{89.52} \\
0.8 & 79.04 & 88.42 \\
1 & 79.93 & 88.66 \\
\bottomrule
\end{tabular}
\end{minipage}
\end{table}

\begin{table}[h]
\caption{Ablation study about number of property text descriptions, LLM is GPT-o1 and loss percentage $\alpha$=0.5.}
\centering
\begin{tabular}{cccccccc}
\hline
&0 &1 &2 &3 &4 &5 &10 \\
\hline
mIoU($\%$)  &79.58 &79.79 &79.93 &80.19 &79.49 &\textbf{80.40} &80.25\\
FB-IoU ($\%$) &89.03 &89.18 &89.27 &89.39 &89.02 &\textbf{89.52} &89.47 \\
\hline
\label{tab_ablation_number}
\end{tabular}
\end{table}

\textbf{Ablation Study on different LLMs.}
The attribute description capability of large language models (LLMs) varies significantly depending on model scale, training methodology, and other architectural factors. As illustrated in Table~\ref{tab:llm_ablation}, our comparative analysis of different LLMs reveals that Qwen-2.5-7B achieves superior performance in generating attribute descriptions, attaining an accuracy of 80.72$\%$ and outperforming other evaluated models. Due to the little performance difference, we used GPT-o1 for all the experiments for speed and convenience considerations.

\textbf{Ablation Study on support images.}
In order to verify our view that support images are difficult to provide effective reference information, we directly remove the support images in Table~\ref{tab:supp} and find that there is a decrease of 0.7$\%$ in VOC, but an increase of 0.9$\%$ in COCO, which suggests that the support images provide a little reference information in the simple dataset, in the complex dataset, due to more significant visual differences, resulting in the support image difficult to provide a reference, even a negative reference.

\begin{table}[!t]
\centering
\begin{minipage}{0.5\textwidth}
\centering
\captionof{table}{Ablation study of different LLMs on different model scale, architecture, \emph{etc.}}
\label{tab:llm_ablation}
\begin{tabular}{ccc}
\toprule
LLMs & mIoU (\%) & FB-IoU (\%) \\
\hline
GPT-o1 & 80.40 & 89.52 \\
GPT-4o & 78.71 & 88.53 \\
DeepSeek-R1 & 79.96 & 86.29 \\
Qwen2.5-3B & 80.39 & 89.60 \\
Qwen2.5-7B & \textbf{80.72} & \textbf{89.80} \\
Qwen2.5-32B & 80.32 & 89.62 \\
\bottomrule
\end{tabular}
\end{minipage}
\hfill
\begin{minipage}{0.5\textwidth}
\centering
\captionof{table}{Ablation study on support images.}
\label{tab:supp}
\begin{tabular}{@{}lcc@{\hspace{0.5em}}cc@{}}
\toprule
 & \multicolumn{2}{c}{PASCAL-5$^i$} & \multicolumn{2}{c}{COCO-20$^i$} \\
\cmidrule(lr){2-3} \cmidrule(lr){4-5}
 & \emph{w/o} supp & \emph{w} supp & \emph{w/o} supp & \emph{w} supp \\
\midrule
fold0 & 79.8 & \textbf{80.4} & \textbf{55.8} & 55.4 \\
fold1 & 84.4 & \textbf{85.4} & \textbf{65.9} & 64.9 \\
fold2 & 74.3 & \textbf{74.8} & \textbf{62.4} & 60.8 \\
fold3 & 74.4 & \textbf{75.6} & \textbf{61.3} & 60.8 \\
\midrule
Mean & 78.3 & \textbf{79.0} & \textbf{61.4} & 60.5 \\
\bottomrule
\end{tabular}
\end{minipage}
\end{table}

\section{Conclusion}\label{sec5}
This paper rethinks FSS support mechanisms, showing the key lies not in support images but in providing complete reference information for query matching. In contrast to visual data, textual data exhibits robustness. Hence, we first introduce the MaE module, which leverages attribute text as an effective reference instead of support images, to generate multiple attribute prior masks for guiding segmentation. Additionally, to bridge the text-visual gap, we propose the MaA module, which aligns attribute textual descriptions with global visual features, yielding enriched visual features that streamline the decoding process. Our experiments reveal that the model can achieve remarkable results even without utilizing support images.

\section{Acknowledgment}\label{sec5}
\textbf{Conflict of interest statement:}
The authors declare that they have no competing interests.\\
\textbf{Compliance of ethical standard statement:}
The study is based on publicly available datasets and synthetic text descriptions generated by AI models.
No human participants or sensitive personal information are involved; therefore, ethical approval is not applicable.\\
\textbf{Informed consent:}
The study relies solely on publicly available datasets. No human subjects or personally identifiable data were involved; hence, informed consent was not applicable.\\
\textbf{Funding Information:}
This work was supported in part by the National Natural Science Foundation of China (Grant No. 62372468, 62301613), in part by the Shandong Natural Science Foundation (Grant No. ZR2023QF046, ZR2023MF008), in part by the Major Basic Research Projects in Shandong Province (Grant No.ZR2023ZD32), in part by the Taishan Scholar Program of Shandong (No. tsqn202306130).\\
\textbf{Data availability:}
The datasets (COCO-20$i$, PASCAL-5$i$) used in this study are publicly available. For other text descriptions generated using large language models, we will open-source them alongside the code after the paper is accepted.



\end{document}